\documentclass[11pt,a4paper]{article}
\usepackage[hyperref]{acl2020}
\usepackage{times}
\usepackage{latexsym}
\usepackage{url}
\usepackage{paralist}
\usepackage{multirow}
\usepackage{verbatim}
\usepackage{amsfonts}
\usepackage{amssymb}
\usepackage{amsmath}
\usepackage{algorithm}
\usepackage{algpseudocode}
\usepackage{graphics}
\usepackage{epsfig}
\usepackage{multicol}
\usepackage{makecell}

\DeclareMathOperator*{\argmax}{arg\,max}

\usepackage{microtype}

\aclfinalcopy 
\def\aclpaperid{3322} 

\title{Towards Unsupervised Language Understanding and Generation\\ by Joint Dual Learning} 

\author{Shang-Yu Su \quad Chao-Wei Huang \quad Yun-Nung Chen \\
Department of Computer Science and Information Engineering\\
National Taiwan University\\
\texttt{ \{f05921117,r07922069\}@ntu.edu.tw\quad y.v.chen@ieee.org }}

\date{}

\begin{document}
\maketitle
\begin{abstract}
In modular dialogue systems, natural language understanding (NLU) and natural language generation (NLG) are two critical components, where NLU extracts the semantics from the given texts and NLG is to construct corresponding natural language sentences based on the input semantic representations.
However, the dual property between understanding and generation has been rarely explored.
The prior work~\cite{su2019dual} is the first attempt that utilized the duality between NLU and NLG to improve the performance via a dual supervised learning framework.
However, the prior work still learned both components in a \emph{supervised} manner, instead,
this paper introduces a general learning framework to effectively exploit such duality, providing flexibility of incorporating both \emph{supervised} and \emph{unsupervised} learning algorithms to train language understanding and generation models in a joint fashion. 
The benchmark experiments demonstrate that the proposed approach is capable of boosting the performance of both NLU and NLG.\footnote{The source code is available at: \url{https://github.com/MiuLab/DuaLUG}.}
\end{abstract}

\section{Introduction}
Spoken dialogue systems that assist users to solve complex tasks such as booking a movie ticket have become an emerging research topic in artificial intelligence and natural language processing areas. 
With a well-designed dialogue system as an intelligent personal assistant, people can accomplish certain tasks more easily via natural language interactions. 
Nowadays, there are several virtual intelligent assistants, such as Apple's Siri, Google Assistant, Microsoft's Cortana, and Amazon's Alexa. 

The recent advance of deep learning has inspired many applications of neural dialogue systems~\cite{wen2017network,bordes2017learning}.
A typical dialogue system pipeline can be divided into several components: a speech recognizer that transcribes a user's speech input into texts, a natural language understanding module (NLU) to classify the domain along with domain-specific intents and fill in a set of slots to form a semantic frame~\cite{tur2011spoken,hakkani2016multi}. 
A dialogue state tracking (DST) module predicts the current dialogue state according to the multi-turn conversations, then the dialogue policy determines the system action for the next turn given the current dialogue state~\cite{peng2018deep,su2018discriminative}.
Finally, the semantic frame indicating the policy is fed into a natural language generationt (NLG) module to construct a response utterance to the user~\cite{wen2015semantically,su2018natural}. 

Generally, NLU is to extract core semantic concepts from the given utterances, while NLG is to construct corresponding sentences based on the given semantic representations.
However, the dual property between understanding and generation has been rarely investigated, \citet{su2019dual} first introduced the duality into the typical supervised learning schemes to train these two models.
Different from the prior work, this paper proposes a general learning framework leveraging the duality between understanding and generation, providing flexibility of incorporating not only \emph{supervised} but also \emph{unsupervised} learning algorithms to jointly train NLU and NLG modules. 
The contributions can be summarized as 3-fold:
\begin{compactitem}
\item This paper proposes a general learning framework using the duality between NLU and NLG, where \emph{supervised} and \emph{unsupervised} learning can be flexibly incorporated for joint training.
\item This work is the first attempt to exploits the dual relationship between NLU and NLG towards unsupervised learning. 
\item The benchmark experiments demonstrate the effectiveness of the proposed framework.
\end{compactitem}

\section{Related Work}
This paper focuses on modeling the duality between understanding and generation towards unsupervised learning of the two components, related work is summarized below.

\paragraph{Natural Language Understanding}
In dialogue systems, the first component is a natural language understanding (NLU) module---parsing user utterances into semantic frames that capture the core meaning~\cite{tur2011spoken}.
A typical NLU first determines the domain given input utterances, predicts the intent, and then fill the associated slots~\cite{hakkani2016multi,chen2016knowledge}.
However, the above work focused on single-turn interactions, where each utterance is treated independently.
To overcome the error propagation and further improve understanding performance, contextual information has been leveraged and shown useful~\cite{chen2015leveraging,sun2016an,shi2015contextual,weston2015memory}.
Also, different speaker roles provided informative signal for capturing speaking behaviors and achieving better understanding performance~\cite{chen2017dynamic,su2018how}.

\paragraph{Natural Language Generation}
NLG is another key component in dialogue systems, where the goal is to generate natural language sentences conditioned on the given semantics from the dialogue manager.
As an endpoint of interacting with users, the quality of generated sentences is crucial for better user experience. 
In spite of robustness and adequacy of the rule-based methods, poor diversity makes talking to a template-based machine unsatisfactory.
Furthermore, scalability is an issue, because designing sophisticated rules for a specific domain is time-consuming.
Previous work proposed a RNNLM-based NLG that can be trained on any corpus of dialogue act-utterance pairs without hand-crafted features and any semantic alignment~\cite{wen2015stochastic}. 
The following work based on sequence-to-sequence (seq2seq) models further obtained better performance by employing encoder-decoder structure with linguistic knowledge such as syntax trees~\cite{sutskever2014sequence,su2018natural}. 

\begin{figure*}[t!]
\centering 
\includegraphics[width=0.9\linewidth]{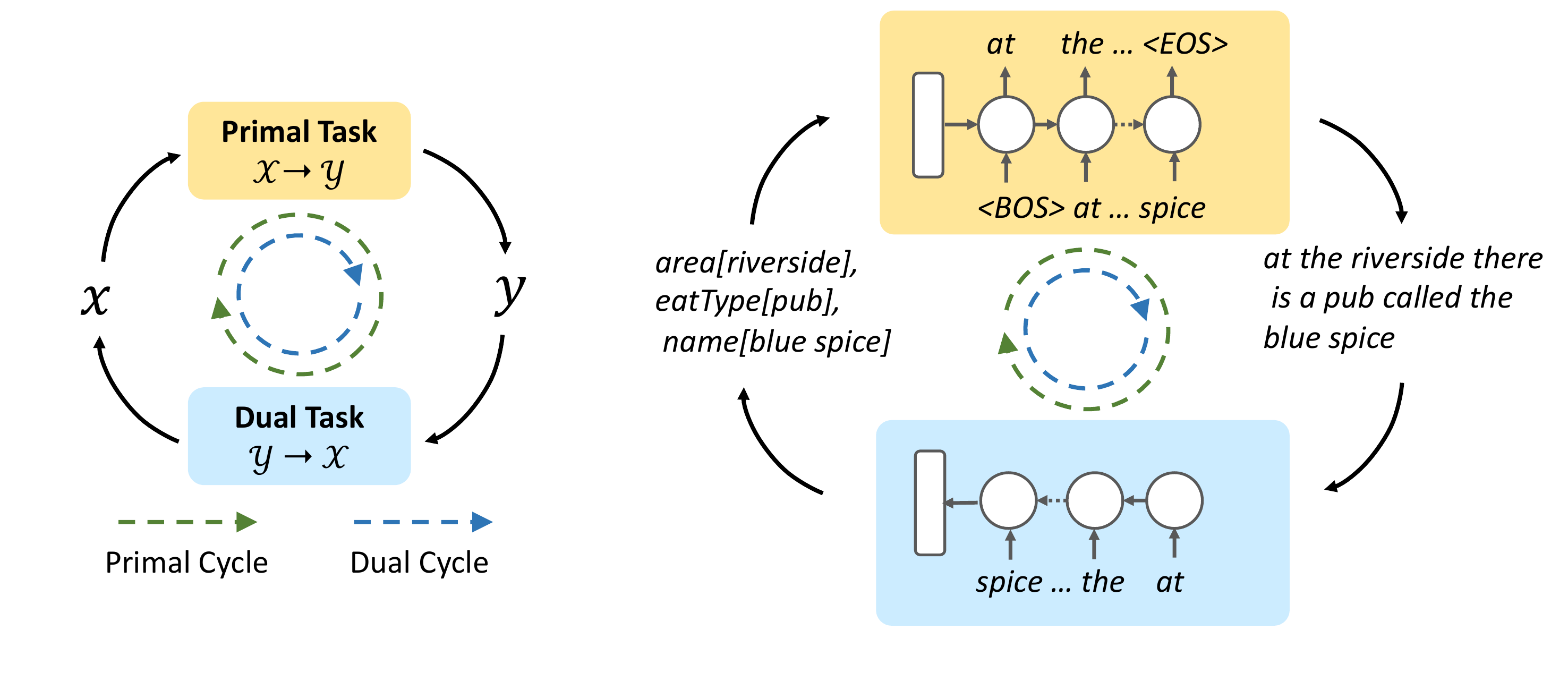}
 \vspace{-5mm}
\caption{\textbf{Left}: The proposed joint dual learning framework, which comprises \emph{Primal Cycle} and \emph{Dual Cycle}. The framework is agnostic to learning objectives and the algorithm is detailed in Algorithm \ref{alg::jdl}. \textbf{Right}: In our experiments, the models for NLG and NLU are a GRU unit accompanied with a fully-connected layer.} 
\label{fig:framework} 
\end{figure*}

\paragraph{Dual Learning}
Various tasks may have diverse goals, which are usually independent to each other.
However, some tasks may hold a dual form, that is, we can swap the input and target of a task to formulate another task.
Such structural duality emerges as one of the important relationship for further investigation.
Two AI tasks are of structure duality if the goal of one task is to learn a function mapping from space $\mathcal{X}$ to $\mathcal{Y}$, while the other’s goal is to learn a reverse mapping from $\mathcal{Y}$ and $\mathcal{X}$.
Machine translation is an example~\cite{wu2016google}, translation from English to Chinese has a dual task, which is translated from Chinese to English; the goal of automatic speech recognition (ASR) is opposite to the one of text-to-speech (TTS)~\cite{tjandra2017listening}, and so on. 
Previous work first exploited the duality of the task pairs and proposed supervised \cite{xia2017dual} and unsupervised (reinforcement learning) \cite{he2016dual} learning frameworks. 
These recent studies magnified the importance of the duality by revealing exploitation of it could boost the learning of both tasks.
\citet{su2019dual} employed the dual supervised learning framework to train NLU and NLG and improve both models simultaneously. 
Recently, \citet{shen-etal-2019-pragmatically} improved models for conditional text generation using techniques from computational pragmatics.
The techniques formulated language production as a game between speakers and listeners, where a speaker should generate text which a listener can use to correctly identify the original input the text describes.

However, although the duality has been considered into the learning objective, two models in previous work are still trained \emph{separately}.
In contrast, this work proposes a general learning framework that trains the models jointly, so that unsupervised learning methods in this research field can be better explored.
\section{Proposed Framework}
In this section, we describe the problem formulation and the proposed learning framework, which is illustrated in Figure~\ref{fig:framework}.

\subsection{Problem Formulation}
The problems we aim to solve are NLU and NLG; for both tasks, there are two spaces: the semantics space $\mathcal{X}$ and the natural language space $\mathcal{Y}$.
NLG is to generate sentences associated with the given semantics, where the goal is to learn a mapping function $f: \mathcal{X} \to \mathcal{Y}$ that transforms semantic representations into natural language.
On the other hand, NLU is to capture the core meaning of sentences, where the goal is to find a function $g: \mathcal{Y} \to \mathcal{X}$ that predicts semantic representations from the given natural language. 

Given $n$ data pairs $\{(x_i, y_i)\}^n_{i=1}$ $i.i.d.$ sampled from the joint space $\mathcal{X} \times \mathcal{Y}$.
A typical strategy for the optimization problem is based on maximum likelihood estimation (MLE) of the parameterized conditional distribution by the trainable parameters $\theta_{x \to y}$ and $\theta_{y \to x}$ as below:
\begin{align*}
f(x;\theta_{x \to y}) = \argmax_{\theta_{x \to y}} P(y \mid x ; \theta_{x \to y} ), \\
g(y;\theta_{y \to x}) = \argmax_{\theta_{y \to x}} P(x \mid y ; \theta_{y \to x} ). 
\end{align*}

The E2E NLG challenge dataset~\cite{novikova2017e2e}\footnote{\url{http://www.macs.hw.ac.uk/InteractionLab/E2E/}} is adopted in our experiments, which is a crowd-sourced dataset of 50k instances in the restaurant domain. 
Each instance is a pair of a semantic frame containing specific slots and corresponding values and a associated natural language utterance with the given semantics.
For example, a semantic frame with the slot-value pairs ``\textsf{name[Bibimbap House], food[English], priceRange[moderate], area [riverside], near [Clare Hall]}'' corresponds to the target sentence ``\emph{Bibimbap House is a moderately priced restaurant who's main cuisine is English food. You will find this local gem near Clare Hall in the Riverside area.}''. 
Although the original dataset is for NLG, of which the goal is to generate sentences based on the given slot-value pairs, we further formulate the NLU task as predicting slot-value pair based on the utterances, which can be viewed as a multi-label classification problem and each possible slot-value pair is treated as an individual label.
The formulation is similar to the prior work~\cite{su2019dual}.

\subsection{Joint Dual Learning}

\begin{algorithm*}[t!]
\small
  \caption{Joint dual learning algorithm}
  \label{alg::jdl}
  \begin{algorithmic}[1]
    \State \textbf{Input}: a mini-batch of $n$ data pairs $\{(x_i, y_i)\}^n_{i=1}$, the function of the primal task $f$, the function of the dual task $g$, the loss function for the primal task $l_{1}(.)$, the loss function for the dual task $l_{2}(.)$, and the learning rates $\gamma_{1}$, $\gamma_{2}$;
    \Repeat
      \State Start from data $x$, transform $x$ by function $f$: $f(x_{i};\theta_{x \to y})$; \Comment Primal Cycle
      \State Compute the loss by $l_{1}(.)$;
      \State Transform the output of the primal task by function $g$: $g(f(x_{i};\theta_{x \to y});\theta_{y \to x})$;
      \State Compute the loss by $l_{2}(.)$;
      \State Update model parameters:
      \State $\theta_{x \to y}$ $\leftarrow$ $\theta_{x \to y}$ - $\gamma_{1} \nabla_{\theta_{x \to y}}(\sum^{n}_{i=1}[l_{1}(f(x_{i};\theta_{x \to y})) + l_{2}(g(f(x_{i};\theta_{x \to y});\theta_{y \to x}))])$;
      \State $\theta_{y \to x}$ $\leftarrow$ $\theta_{y \to x}$ - $\gamma_{2} \nabla_{\theta_{y \to x}} (\sum^{n}_{i=1}[l_{2}(g(f(x_{i};\theta_{x \to y});\theta_{y \to x}))])$;
      \State Start from data $y$, transform $y$ by function $g$: $g(y_{i};\theta_{y \to x})$; \Comment Dual Cycle
      \State Compute the loss by $l_{2}(.)$;
      \State Transform the output of the dual task by function $f$: $f(g(y_{i};\theta_{y \to x});\theta_{x \to y})$;
      \State Compute the loss by $l_{1}(.)$;
      \State Update model parameters:
      \State $\theta_{y \to x}$ $\leftarrow$ $\theta_{y \to x}$ - $\gamma_{2} \nabla_{\theta_{y \to x}} (\sum^{n}_{i=1}[l_{2}(g(y_{i};\theta_{y \to x})) + l_{1}(f(g(y_{i};\theta_{y \to x});\theta_{x \to y}))])$;
      \State $\theta_{x \to y}$ $\leftarrow$ $\theta_{x \to y}$ - $\gamma_{1} \nabla_{\theta_{x \to y}} (\sum^{n}_{i=1}[l_{1}(f(g(y_{i};\theta_{y \to x});\theta_{x \to y}))])$;
    \Until{convergence}
  \end{algorithmic}
\end{algorithm*}

Although previous work has introduced the learning schemes that exploit duality of AI tasks, most of it was based on reinforcement learning or standard supervised learning and the models of primal and dual tasks ($f$ and $g$ respectively) are trained \emph{separately}.
Intuitively, if the models of primal and dual tasks are optimally learned, a complete cycle of transforming data from the original space to another space then back to the original space should be exactly the same as the original data, which could be viewed as the ultimate goal of a dual problem.
In our scenario, if we generate sentences from given semantics $x$ via the function $f$ and transform them back to the original semantics perfectly via the function $g$, it implies that our generated sentences are grounded to the original given semantics and has the mathematical condition:
\begin{align*}
g(f(x)) \equiv x.
\end{align*}
Therefore, our objective is to achieve the \emph{perfect complete cycle} of data transforming by training two dual models ($f$ and $g$) in a \emph{joint} manner.

\subsubsection{Algorithm Description}
As illustrated in Figure \ref{fig:framework}, the framework is composed of two parts: \emph{Primal Cycle} and \emph{Dual Cycle}.
Primal Cycle starts from semantic frames $x$, 
(1) first transforms the semantic representation to sentences by the function $f$, 
(2) then computes the loss by the given loss function $l_{1}$, 
(3) predicts the semantic meaning from the generated sentences, 
(4) computes the loss by the given loss function $l_{2}$, 
(5) finally train the models based on the computed loss;
Dual Cycle starts from utterances and is symmetrically formulated. 
The learning algorithm is described in Algorithm \ref{alg::jdl}, which is agnostic to types of learning objective.
Either a supervised learning objective or an unsupervised learning objective can be conducted at the end of the training cycles, and the whole framework can be trained in an \emph{end-to-end} manner.

\subsection{Learning Objective}
As the language understanding task in our experiments is to predict corresponding slot-value pairs of utterances, which is a multi-label classification problem, we utilized the binary cross entropy loss as the supervised objective function for NLU. 
Likewise, the cross entropy loss function is used as the supervised objective for NLG.
Take NLG for example, the objective of the model is to optimize the conditional probability of predicting word tokens given semantics $p(y \mid x)$, so that the difference between the predicted distribution and the target distribution, $q(y \mid x)$, can be minimized:
\begin{equation}
\label{eq:ce}
-\sum^{n}\sum_{y}q(y \mid x) \log p(y \mid x),
\end{equation}
where $n$ is the number of samples.

On the other hand, we can also introduce the reinforcement learning objective into our framework, the objective aims to maximize the expected value of accumulated reward.
In our experiments, we conduct policy gradient (REINFORCE) method \cite{sutton2000policy} for optimization, the gradient could be written as: 
\begin{align}
\label{eq:pg}
\nabla \mathbb{E}[r] = \mathbb{E}[r(y)\nabla \log p(y \mid x)],
\end{align}
where the variety of reward $r$ will be elaborated in the next section.  
The loss function $l_{1}$ for both tasks could be  (\ref{eq:ce}), (\ref{eq:pg}), and the combination of them. 

\subsection{Reward Function}
Different types of rewards reflect various objectives and would result in different behaviors in the learned policy.
Hence, we design various reward functions to explore the model behavior, including explicit and implicit feedback.

\subsubsection{Explicit Reward}
To evaluate the quality of generated sentences, two explicit reward functions are adopted.
\paragraph{Reconstruction Likelihood}
In our scenario, if we generate sentences based on given semantics $x$ by the function $f$ and could transform them back to the original semantics perfectly by the function $g$, it implies our generated sentences ground on the original given semantics.
Therefore we use the reconstruction likelihood at the end of the training cycles as a reward function:
\begin{align*}
 \begin{cases}
  \log p(x \mid f(x_{i};\theta_{x \to y});\theta_{y \to x}) & \textbf{Primal},\\
  \log p(y \mid g(y_{i};\theta_{y \to x});\theta_{x \to y}) & \textbf{Dual}. \\
 \end{cases}
\end{align*}

\paragraph{Automatic Evaluation Score}
The goal of most NLP tasks is to predict word tokens correctly, so the loss functions used to train these models focus on the word level, such as cross entropy maximizing the continuous probability distribution of the next correct word given the preceding context. 
However, the performance of these models is typically evaluated using discrete metrics.
For instance, BLEU and ROUGE measure n-gram overlaps between the generated outputs and the reference texts.
In order to enforce our NLG to generate better results in terms of the evaluation metrics, we  utilize these automatic metrics as rewards to provide the sentence-level information.
Moreover, we also leverge F-score in our NLU model to indicate the understanding performance.

\subsubsection{Implicit Reward}
In addition to explicit signals like reconstruction likelihood and the automatic evaluation metrics, a ``softer'' feedback signal may be informative.
For both tasks, we design model-based methods estimating data distribution in order to provide such soft feedback.

\paragraph{Language Model}
For NLG, we utilize pre-trained language models which estimate the whole data distribution to compute the joint probability of generated sentences, measuring their naturalness and fluency.
In this work, we use a simple language model based on RNN~\cite{mikolov2010recurrent,sundermeyer2012lstm}. 
The language model is learned by a cross entropy objective in an unsupervised manner:
\begin{align}
p(y) = \prod_{i}^{L} p(y_{i} \mid y_{1}, ... , y_{i-1}; \theta_{y}),
\label{eq:lm}
\end{align}
where $y_{(\cdot)}$ are the words in a sentence $y$, and $L$ is the length of the utterance.  

\paragraph{Masked Autoencoder for Distribution Estimation (MADE)}
For NLU, the output contains a set of discrete labels, which do not fit the sequential model scenarios such as language models.
Each semantic frame $x$ in our work contains the core concept of a certain sentence, furthermore, the slot-value pairs are not independent to others, because they correspond to the same individual utterance.
For example, McDonald's would probably be inexpensive;
therefore the correlation should be taken into account when estimating the joint distribution.

Following \citet{su2019dual}, we measure the soft feedback signal for NLU using masked autoencoder~\cite{germain2015made} to estimate the joint distribution.
By interrupting certain connections between hidden layers, we could enforce the variable unit $x_{d}$ to only depend on any specific set of variables, not necessary on $x_{<d}$; eventually we could still have the joint distribution by product rule:
\begin{equation*}
p(x) = \prod_{d}^{D} p(x_{d} \mid S_{d} ),
\end{equation*}
where $d$ is the index of variable unit, $D$ is the total number of variables, and $S_{d}$ is a specific set of variable units.
Because there is no explicit rule specifying the exact dependencies between slot-value pairs in our data, we consider various dependencies by ensembles of multiple decomposition by sampling different sets $S_{d}$ and averaging the results.

\subsection{Flexibility of Learning Scheme}
The proposed framework provides various flexibility of designing and extending the learning scheme, described as follows.

\paragraph{Straight-Through Estimator}
In many NLP tasks, the learning targets are discrete, so the goals of most NLP tasks are predicting discrete labels such as words.
In practice we perform \texttt{argmax} operations on the output distribution from learned models to select the most possible candidates.
However, such operation does not have any gradient value, forbidding the networks be trained via backpropagation.
Therefore, it is difficult to directly connect a primal task (NLU in our scenario) and a dual task (NLG in our scenario) and jointly train these two models due to the above issue. 

The Straight-Through (ST) estimator~\cite{bengio2013estimating} is a widely applied method due to its simplicity and effectiveness.
The idea of Straight-Through estimator is directly using the gradients of discrete samples as the gradients of the distribution parameters.
Because discrete samples could be generated as the output of hard threshold functions or some operations on the continuous distribution, \citet{bengio2013estimating} explained the estimator by setting the gradients of hard threshold functions to 1.
In this work, we introduce ST estimator for connecting two models, and therefore the gradient can be estimated and two models can be jointly trained in an end-to-end manner.

\paragraph{Distribution as Input}
In addition to employing the Straight-Through estimator, an alternative solution is to use continuous distribution as the input of models. 
For NLU, the inputs are the word tokens from NLG, so we use the predicted distribution over the vocabulary to perform the weighted-sum of word embeddings. 
For NLG, the model requires semantic frame vectors predicted by NLU as the input condition; in this case, the probability distribution of slot-value pairs predicted by NLU can directly serve as the input vector.   
By utilizing the output distribution in this way, two models can be trained jointly in an end-to-end fashion. 


\paragraph{Hybrid Objective}
As described before, the proposed approach is agnostic to learning algorithms; in other words, we could apply different learning algorithms at the middle and end of the cycles.
For example, we could apply supervised learning on NLU in the first half of Primal Cycle and reinforcement learning on NLG to form a hybrid training cycle.
Because two models are trained jointly, the objective applied on one model would potentially impact on the behavior of the other.
Furthermore, we could also apply multiple objective functions including supervised or unsupervised ones to formulate multi-objective learning schemes.

\paragraph{Towards Unsupervised Learning}
Because the whole framework can be trained jointly and propagate the gradients, we could apply only one objective in one learning cycle at the end of it. 
Specifically, in Algorithm \ref{alg::jdl}, we can apply only $l_{2}$ in line 8 and only $l_{1}$ in line 15.
Such flexibility potentially enables us to train the models based on unpaired data in a unsupervised manner.
For example, sample unpaired data $x$ and transform the data by function $f$, next, feed them into the function $g$, then compare the predicted results and the original input to compute the loss. 
Likewise, we can perform the training cycle symmetrically from $y$.
It is also possible to utilize limited data and perform the autoencoding cycle described above to apply semi-supervised learning.

\begin{table*}[t!]
\centering
\begin{tabular}{ | c| l | c | c c c c| }
    \hline
    \multicolumn{2}{|c|}{\multirow{2}{*}{\bf Learning Scheme}} & \bf NLU & \multicolumn{4}{c|}{\bf NLG} \\
    \multicolumn{2}{|c|}{} & \bf \small Micro-F1 & \bf \small BLEU & \bf \small ROUGE-1 & \bf \small ROUGE-2 & \bf \small ROUGE-L  \\
\hline \hline
(a) & Iterative training (\textbf{supervised}) &  71.14 & 55.05 & 55.37 & 27.95 & 39.90 \\
(b) & Dual supervised learning \cite{su2019dual} & \bf 72.32 & \bf 57.16 & \bf 56.37 & \bf 29.19 & \bf 40.44\\
\hline
(c) & Joint training (Straight-Through) & 71.73 & 55.19 & 55.16 & 27.45 & 39.33 \\
(d) & (c) + (NLG w/ distribution) & 73.22 & 55.18 & 55.35 & 27.81 & \bf 39.36 \\
(e) & (c) + (NLU w/ distribution) & 79.19 & 51.47 & 53.62 & 26.17 & 37.90 \\
(f) & (c) + (NLU and NLG w/ distribution) & \bf 80.03 & \bf 55.34 & \bf 56.17 & \bf 28.48 & 39.24 \\
\hline
(g) & (f) + \textbf{RL}$_{mid}$(\text{reconstruction likelihood}) & 80.07 & 55.32 & 56.12 & 28.07 & 39.59 \\
(h) & (f) + \textbf{RL}$_{end}$(\text{reconstruction likelihood}) & 79.97 & 55.21 & 56.15 & 28.50 & 39.42 \\
(i) & (f) + \textbf{RL}$_{mid}$(\text{BLEU+ROUGE, F1}) & 79.49 & 56.04 & 56.61 & 28.78 & 39.93 \\
(j) & (f) + \textbf{RL}$_{end}$(\text{BLEU+ROUGE, F1}) & 80.35 &  \bf 57.59 & \bf  56.71 & \bf  29.06 & \bf  40.28 \\
(k) & (f) + \textbf{RL}$_{mid}$(\text{LM, MADE}) & \bf 81.52 & 54.13 & 54.60 & 26.85 & 38.90 \\
(l) & (f) + \textbf{RL}$_{end}$(\text{LM, MADE}) & 79.52 & 55.61 & 55.97 & 28.57 & 39.97 \\
\hline
  \end{tabular}
\vspace{-2mm}
\caption{The NLU performance reported on micro-F1 and the NLG performance reported on BLEU, ROUGE-1, ROUGE-2, and ROUGE-L of models (\%).}
\vspace{-3mm}
\label{tab:results}
\end{table*}

\section{Experiments}

Our models are trained on the official training set and verified on the official testing set of the E2E NLG challenge dataset~\cite{novikova2017e2e}. 
The data preprocessing includes trimming punctuation marks, lemmatization, and turning all words into lowercase. 
Each possible slot-value pair is treated as an individual label and the total number of labels is 79.
To evaluate the quality of the generated sequences regarding both precision and recall, for NLG, the evaluation metrics include BLEU and ROUGE (1, 2, L) scores with multiple references, while F1 measure is reported for evaluating NLU.

\subsection{Model}
The proposed framework and algorithm are agnostic to model structures.
In our experiments, we use a gated recurrent unit (GRU) ~\cite{cho2014learning} with fully-connected layers at ends of GRU for both NLU and NLG, which are illustrated in the right part of Figure \ref{fig:framework}.
Thus the models may have semantic frame representation as initial and final hidden states and sentences as the sequential input.
In all experiments, we use mini-batch \textit{Adam} as the optimizer with each batch of 64 examples.
10 training epochs were performed without early stop, the hidden size of network layers is 200, and word embedding is of size 50. 

\subsection{Results and Analysis}
The experimental results are shown in Table \ref{tab:results}, each reported number is averaged on the official testing set from three turns.
Row (a) is the baseline where NLU and NLG models are trained independently and separately by supervised learning. 
The best performance in \cite{su2019dual} is reported in row (b), where NLU and NLG are trained separately by supervised learning with regularization terms exploiting the duality.

To overcome the issue of non-differentiability, we introduce Straight-Through estimator when connecting two tasks.
Based on our framework, another baseline for comparison is to train two models jointly by supervised loss and straight-through estimators, of which the performance is reported in row (c).
Specifically, the cross entropy loss (\ref{eq:ce}) is utilized in both $l_{1}$ and $l_{2}$ in Algorithm \ref{alg::jdl}.
Because the models in the proposed framework are trained jointly, the gradients are able to flow through the whole network thus two models would directly influence learning of each other. 
Rows (d)-(f) show the ablation experiments for exploring the interaction between two models ($f$ and $g$).
For instance, row (e) does not use ST at the output of the NLU module; instead, we feed continuous distribution over slot-value labels instead of discrete semantic frames into NLG as the input.
Instead of discrete word labels, row (d) and row (f) feed weighted sum over word embeddings based on output distributions.
%
Since the goal of NLU is to learn a many-to-one function, considering all possibility would potentially benefit learning (row (d)-(f)).

On the contrary, the goal of NLG is to learn a one-to-many function, applying the ST estimator at the output of NLU only rather than both sides degrades the performance of generation (row (e)).
However, this model achieves unexpected improvement in understanding by over 10\%, the reason may be the following. 
The semantics representation is very compact, a slight noise in the semantics space would possibly result in a large difference in the target space and a totally different semantic meaning.
Hence the continuous distribution over slot-value pairs may potentially cover the unseen mixture of semantics and further provide rich gradient signals.
This could also be explained from the perspective of data augmentation.
Moreover, connecting two models with continuous distribution at both joints further achieves improvement in both NLU and NLG (row (f)).
Although row (f) performs best in our experiments and dataset, as most AI tasks are classification problems, the proposed framework with ST estimators provides a general way to connect two tasks with duality. 
The proposed methods also significantly outperform the previously proposed dual supervised learning framework \cite{su2019dual} on F1 score of NLU and BLEU score of NLG, demonstrating the benefit of learning NLU and NLG jointly. 

\begin{table*}[t!]
\centering
 \small
\begin{tabular}{ | c  | p{6cm}|p{6cm} | }
\hline
    & \bf Baseline & \bf Proposed\\
    \hline
$x$ & \multicolumn{2}{l|}{area[riverside], eatType[pub], name[blue spice]}\\
    \hline
$y$ & \multicolumn{2}{l|}{ \it at the riverside there is a pub called the blue spice}\\
    \hline
$f(x;\theta_{x \to y})$ & \it blue spice is a pub in riverside that has a price range of more than 30e & \it in riverside there is a pub called blue spice\\
    \hline
$g(f(x;\theta_{x \to y});\theta_{y \to x}))$ & area[city centre], customer rating[5 out of 5], priceRange[more than 30], priceRange[cheap], name[blue spice], name[the vaults] & area[riverside], eatType[pub], name[blue spice]\\
    \hline
  \end{tabular}
\vspace{-2mm}
\caption{An example of Primal Cycle, where the baseline model is row (a) in Table \ref{tab:results}.}
\vspace{-2mm}
\label{tab:examples-SNS}
\end{table*}

\begin{table*}[t!]
\centering
 \small
\begin{tabular}{ | c | p{6cm}| p{6cm} | }
    \hline
    & \bf Baseline & \bf Proposed \\
    \hline
 $y$ & \multicolumn{2}{p{12cm}|}{\it blue spice is a family friendly pub located in the city  centre it serves chinese food and is near the rainbow vegetarian cafe} \\
    \hline
 $x$ & \multicolumn{2}{p{12cm}|}{familyFriendly[yes], area[city centre], eatType[pub], food[chinese], name[blue spice], near[rainbow vegetarian cafe]}\\
    \hline
 $g(y;\theta_{y \to x}))$ & familyFriendly[yes], food:[chinese] & familyFriendly[yes], area[city centre], eatType[pub], priceRange[moderate], food[chinese], name[blue spice]\\
    \hline
 $f(g(y;\theta_{y \to x}));\theta_{x \to y})$ & \it the chinese restaurant the twenty two is a family friendly restaurant & \it the chinese restaurant the blue spice is located in the city centre it is moderately priced and kid friendly\\
 \hline
  \end{tabular}
\vspace{-2mm}
\caption{An example of Dual Cycle, where the baseline model is row (a) in Table \ref{tab:results}.}
\vspace{-3mm}
\label{tab:examples-NSN}
\end{table*}

\subsection{Investigation of Hybrid Objectives}
The proposed framework provides the flexibility of applying multiple objectives and different types of learning methods.
In our experiments, apart from training two models jointly by supervised loss, reinforcement learning objectives are also incorporated into the training schemes (row (g)-(l)).
The ultimate goal of reinforcement learning is to maximize the expected reward (equation (\ref{eq:pg})).
In the proposed dual framework, if we take expectation over different distribution, it would reflect a different physical meaning.
For instance, if we receive a reward at the end of Primal Cycle and the expectation is taken over the output distribution of NLG (middle) or NLU (end), the derivatives of objective functions would differ:
\begin{align*}
\label{eq:a}
\begin{cases}
\mathbb{E}[r_{i}\nabla \log p(y_i \mid x ;\theta_{x \to y})] & \textbf{RL}_{mid}, \\
\mathbb{E}[r_{i}\nabla \log p(x_i \mid f(x;\theta_{x \to y}) ;\theta_{y \to x})] & \textbf{RL}_{end}. 
\end{cases}
\end{align*}
The upper one ($\textbf{RL}_{mid}$) assesses the expected reward earned by the sentences constructed by the policy of NLG, which is a direct signal for the primal task NLG. 
The lower one ($\textbf{RL}_{end}$) estimates the expected reward earned by the predicted semantics by the policy of NLU based on the state predicted by NLG, such reward is another type of feedback.

In the proposed framework, the models of two tasks are trained jointly, thus an objective function will simultaneously influence the learning of both models.
Different reward designs could guide reinforcement learning agents to different behaviors. 
To explore the impact of reinforcement learning signal, various rewards are applied on top of the joint framework (row (f)): 
\begin{compactenum}
\item token-level likelihood (rows (g) and (h)), 
\item sentence/frame-level automatic evaluation metrics (rows (i) and (j)), 
\item corpus-level joint distribution estimation (rows (k) and (l)).  
\end{compactenum}
In other words, the models in rows (g)-(l) have both supervised and reinforcement learning signal.
The results show that token-level feedback may not provide extra guidance (rows (g) and (h)), directly optimizing towards the evaluation metrics at the testing phase benefits learning in both tasks and performs best (rows (i) and (j)), and the models utilizing learning-based joint distribution estimation also obtain improvement (row (k)).
In sum, the explicit feedback is more useful for boosting the NLG performance, because the reconstruction and automatic scores directly reflect the generation quality.
However, the implicit feedback is more informative for improving NLU, where MADE captures the salient information for building better NLU models.
The results align well with the finding in \citet{su2019dual}.

\subsection{Qualitative Analysis}
Table \ref{tab:examples-SNS} and \ref{tab:examples-NSN} show the selected examples of the proposed model and the baseline model in Primal and Dual Cycle.
As depicted in Algorithm \ref{alg::jdl}, Primal Cycle is designed to start from semantic frames $x$, then transform the representation by the NLG model $f$, finally feed the generated sentences into the NLU model $g$ and compare the results with the original input to compute loss.
In the example of Primal Cycle (Table \ref{tab:examples-SNS}), we can find that $f(g(y;\theta_{y \to x}));\theta_{x \to y})$ equals $x$, which means the proposed method can successfully restore the original semantics.
On the other hand, Dual Cycle starts from natural language utterances, from the generated results (Table \ref{tab:examples-NSN}) we can find that our proposed method would not lose semantic concepts in the middle of the training cycle ($g(y;\theta_{y \to x})) \leftrightarrow x$).
Based on the qualitative analysis, we can find that by considering the duality into the objective and jointly training, the proposed framework can improve the performance of NLU and NLG simultaneously.

\section{Future Work}
Though theoretically sound and empirically validated, the formulation of the proposed framework depends on the characteristics of data.
Not every NLU dataset is suitable for being used as a NLG task, vice versa.
Moreover, though the proposed framework provides possibility of training the two models in a fully-unsupervised manner, it is found unstable and hard to optimize from our experiments.
Thus, better dual learning algorithms and leveraging pretrained models and other learning techniques like adversarial learning are worth exploring to improve our framework, we leave these in our future work.   

\section{Conclusion}
This paper proposes a general learning framework leveraging the duality between language understanding and generation, providing flexibility of incorporating supervised and unsupervised learning algorithm to jointly train two models. 
Such framework provides a potential method towards unsupervised learning of both language understanding and generation models by considering their data distribution.
The experiments on benchmark dataset demonstrate that the proposed approach is capable of boosting the performance of both NLU and NLG models.

\bibliography{anthology,acl2020}
\bibliographystyle{acl_natbib}

\end{document}